\documentclass[letterpaper]{article} 
\usepackage{aaai23} 
\usepackage{times}  
\usepackage{helvet}  
\usepackage{courier}  
\usepackage[hyphens]{url}  
\usepackage{graphicx} 
\usepackage{natbib}  
\usepackage{caption} 
\usepackage{algorithm}
\usepackage{algorithmic}

\usepackage{url}            
\usepackage{newfloat}
\usepackage{listings}
\DeclareCaptionStyle{ruled}{labelfont=normalfont,labelsep=colon,strut=off} 
\floatstyle{ruled}
\newfloat{listing}{tb}{lst}{}
\floatname{listing}{Listing}
\usepackage{booktabs}
\usepackage{multirow}
\usepackage{xcolor}
\usepackage{amsfonts}
\usepackage{amsmath}
\usepackage{tikz}
\newcommand*\circled[1]{\tikz[baseline=(char.base)]{
            \node[shape=circle,draw,inner sep=1pt] (char) {#1};}}

\newcommand{\wrt}{\textit{w.r.t. }}
\newcommand{\ie}{\textit{i.e. }}
\newcommand{\eg}{\textit{e.g. }}
\newcommand{\etal}{\textit{et al. }}

\def\G{{\textrm G}}
\def\D{{\textrm D}}

\title{Fair Generative Models via Transfer Learning}

\author {
    Christopher T.H.Teo\thanks{Equal Contribution \hspace{2 mm} $^\dag$Corresponding Author},
    Milad Abdollahzadeh\footnotemark[1], 
    Ngai-man Cheung$^\dag$
}
\affiliations {
    Singapore University of Technology and Design (SUTD)\\
    {christopher\_teo@mymail.sutd.edu.sg, \{milad\_abdollahzadeh, ngaiman\_cheung\}@sutd.edu.sg}
}

\begin{document}

\maketitle

\begin{abstract}

This work addresses {\em fair} generative models.
Dataset biases have been a major cause of unfairness in deep generative models.
Previous work had proposed to
augment  {\em large, biased} datasets with {\em small, unbiased} reference datasets.
Under this setup, 
a weakly-supervised approach has been proposed, which achieves state-of-the-art quality and fairness in generated samples.
{\bf In our work}, based on this setup, we propose a simple yet effective approach.
Specifically, first, we propose {\bf fairTL}, 
a {\em transfer learning} approach to learn fair generative models. Under fairTL, we pre-train the generative model with the available large, biased datasets and subsequently adapt the model using the small, unbiased reference dataset. 
Our fairTL can learn expressive sample generation during pre-training, thanks to the large (biased) dataset. This knowledge is then transferred to the target model during adaptation, which also learns to capture the underlying fair distribution of the small reference dataset.
Second, we propose {\bf fairTL++}, where we introduce two additional innovations to improve upon fairTL:
(i) multiple feedback and (ii)  Linear-Probing followed by Fine-Tuning (LP-FT). 
Taking one step further, 
we consider an alternative, challenging setup when only a pre-trained (potentially biased) model is available but the dataset used to pre-train the model is {\em inaccessible}.
We demonstrate that our proposed fairTL and fairTL++ remains very effective under this setup.
We note that previous work requires access to large, biased datasets and cannot handle this more challenging setup.
Extensive experiments show that
fairTL and fairTL++ achieve state-of-the-art in both quality and fairness of generated samples.
{\em The code and additional resources can be found at \url{bearwithchris.github.io/fairTL/}}
\end{abstract}

\section{Introduction}
Deep generative models such as Generative Adversarial Network (GAN) 
are an active research area
\cite{goodfellow2014GAN,zhangSelfAttentionGenerativeAdversarial2019a,brockLargeScaleGAN2019,karrasProgressiveGrowingGANs2018a,karrasAnalyzingImprovingImage2020,ojhaFewshotImageGeneration2021}.
Various GAN-based approaches have achieved outstanding results in many tasks, for example: image synthesis 
\cite{karras-cvpr-2019,yu-tip-2019ea}
, image transformation 
\cite{wang-tip-2018perceptual}
, super-resolution 
\cite{lucas-tip-2019generative,nasrollahi2020deep}
, text-to-image synthesis \cite{zhang2-cvpr-2017}, video captioning \cite{yang-tip-2018video}, and anomaly detection \cite{schlegl-ipmi-2017,lim-icdm-2018}.

In recent times, {\bf fairness in generative models}  has attracted increasing attention~
\cite{frankelFairGenerationPrior2020,choiFairGenerativeModeling2020a,humayunMaGNETUniformSampling2021a,tanImprovingFairnessDeep2020}.
It is defined as the \textit{equal representation} \cite{hutchinson50YearsTest2019} of some selected \textit{sensitive attribute} (SA). For example, a generative model that has an equal probability of producing male and female samples is fair \wrt \texttt{Gender}. 
Generative models have been increasingly adopted in various applications including high-stakes areas such as criminal justice \cite{jalanSuspectFaceGeneration2020} and healthcare~\cite{frid-neurocomputing-2018gan}. This brings about concerns regarding potential biases and unfairness of these models.
For example, generative models have been applied in suspect facial profiling \cite{jalanSuspectFaceGeneration2020}. In this application, a generative model could result in wrongful incrimination of an individual if 
the model has 
biases \wrt 
certain SA
such as \texttt{Gender} or \texttt{Race}. 
Furthermore, some generative models have been applied to create data for training downstream models \eg classifiers for disease diagnosis~\cite{frid-neurocomputing-2018gan}. Such biases in generative models can 
propagate to 
downstream models, exacerbating the situation.

\begin{figure*}[ht]
    \centering
    \includegraphics[width=\textwidth]{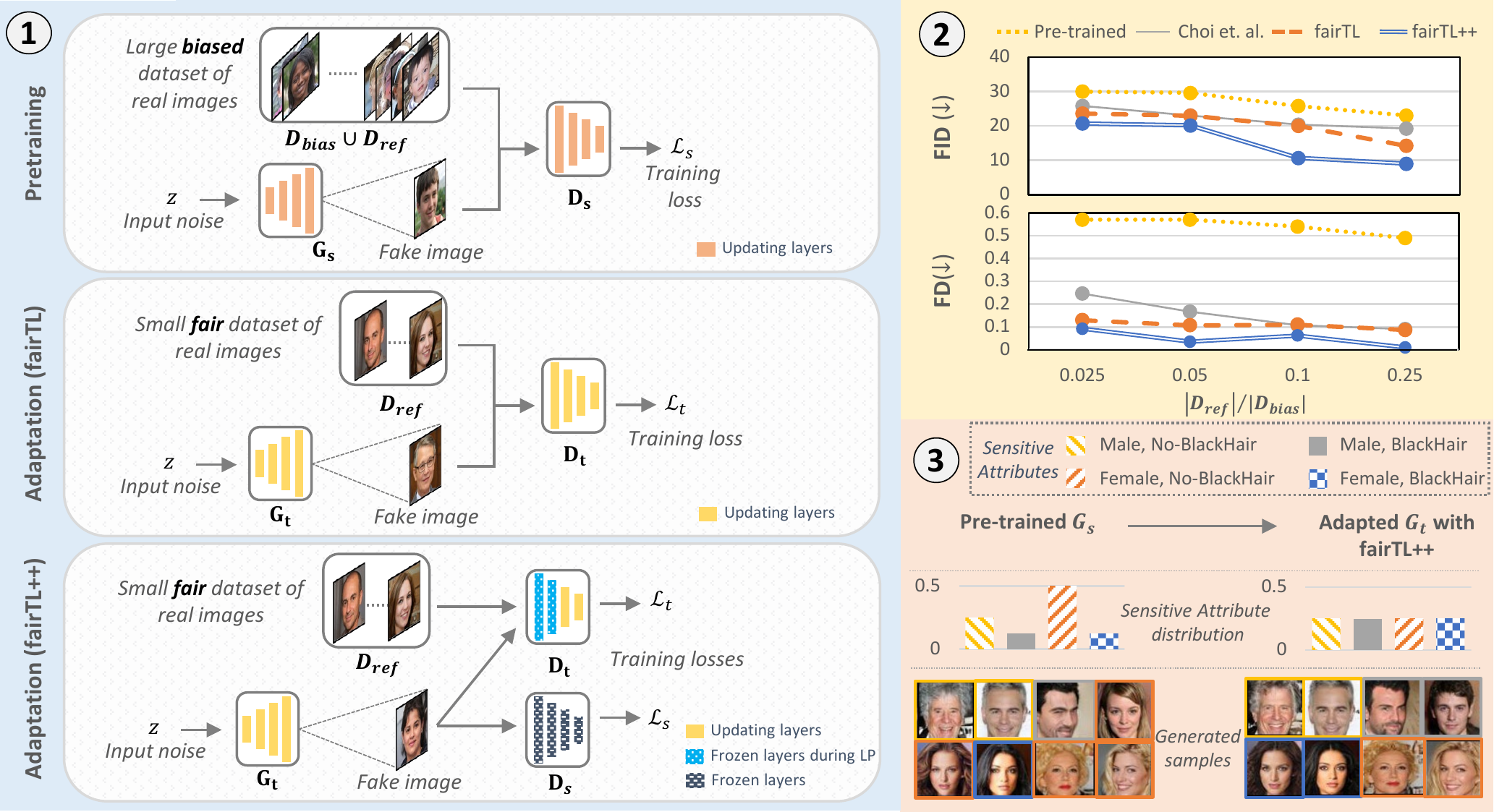}
    \caption{{\bf 
    Overview of our work on training fair generative models.} \circled{1} We train a high-quality generator with fair sensitive attribute (SA) distribution in a two-stage process that we call \textit{fairTL}. 
    In \textit{pre-training}, the GAN learns to generate diverse and high-quality samples from a large but biased dataset.
    Then, in \textit{adaptation}, the same GAN learns the fair underlying SA distribution from a small reference distribution, $D_{ref}$. 
    To improve adaptation, we introduced a second variation called \textit{fairTL++}, which includes an additional source of feedback ($\D_s$) and a Linear-Probing step prior to Fine-Tuning. Pre-training step is same for both fairTL and fairTL++.
    \circled{2} 
    Our results from training a GAN with a large biased dataset, $D_{bias}$, with SA distribution of 90\% females and 10\% males and a small fair dataset, $D_{ref}$, with varying $|D_{ref}|$ denoted by $perc=|D_{ref}|/|D_{bias}|$, from celebA \cite{liu2015faceattributes}. We compare four approaches
    \textit{1) pre-training},
    2) \cite{choiFairGenerativeModeling2020a}: the SOTA technique, 
    3) our proposed \textit{fairTL} and 4) \textit{fairTL++}. We then  measure the Fréchet Inception Distance (FID) and Fairness Discrepancy (FD) of the four models. 
    A smaller FID indicates better quality and smaller FD indicates better fairness.
    Without consideration of fairness, the \textit{pre-trained} setup expresses a large bias. Choi \etal then significantly improves on this but brings about diminishing quality and fairness as 
    $|D_{ref}|$
    becomes smaller. Meanwhile, our proposed method demonstrates greater robustness under these same
    limitations, achieving SOTA results in both FID and FD. 
    \circled{3} We illustrate the improved fairness on multiple SA \texttt{\{Gender,Blackhair\}} during the adaptation stage. To do this, we utilize a fixed noise vector, $z$ to sample from both the pre-trained and fairTL++ models. Observe how the majority-represented SA are adapted to the minority-represented SA, thereby improving the SA distribution. 
    }
    \label{fig:SampleIllustration}
\end{figure*}

{\bf Dataset biases} are a major cause of unfairness in deep generative models. Typically, generative models like GANs are trained in an unsupervised manner to capture the underlying distribution of the given dataset, and then generate new data from the same distribution. 
It is usually expected that the training dataset is large and unbiased \wrt SAs. 
This assumption usually holds true 
when we follow good practices for data collection, such as protocols adopted by biotech companies, or governmental and international institutions such as the World Bank 
\cite{choiFairGenerativeModeling2020a,katalBigDataIssues2013,chizzolaTARGETTakingReflexive2017}.
However, these protocols are usually unscalable, and collected \textit{fair datasets are usually small} in size \cite{choiFairGenerativeModeling2020a}.
Therefore, in order to collect the required large dataset, we usually use alternative sources with a related distribution, such as scrapping images from the internet \cite{muehlethaler2021collecting}.
Collected data from these 
\textit{alternative resources are usually biased} \wrt SAs
\cite{le2022survey,hwangFairFaceGANFairnessawareFacial2020}
, and these biases are easily 
picked up 
by
the generative models.

To prevent the biased dataset 
from harming
the fairness of the generative model,
the large biased dataset $D_{bias}$ can be augmented with a small fair (\wrt some specific SAs) dataset $D_{ref}$, as proposed in
\cite{choiFairGenerativeModeling2020a}.
In this setup, the main idea is that the generative model can learn expressive representation using $D_{bias}$, while mitigating the bias 
with $D_{ref}$.
Note that in their setup, neither datasets are labeled \wrt SAs, and the size of the fair dataset can be much smaller than the biased dataset. For example, $|D_{ref}|$ could be 2.5\% of $|D_{bias}|$.

{\bf In our work}, initially, we follow the setup as in \cite{choiFairGenerativeModeling2020a}, and propose a simple {\em transfer learning} approach for bias mitigation.
Specifically, we first propose fair transfer learning (\textbf{fairTL}) where we pre-train the generative model using the large biased dataset to learn expressive sample generation. Subsequently, on top of the learned expressive knowledge, we adapt the model using the small fair dataset to capture the fair SA adaptation.
We show that this simple transfer learning approach can be considered as a {\em strong baseline} for training a fair generative model via transfer learning (Figure \ref{fig:SampleIllustration}).
However, as $D_{ref}$ is small, the fine-tuning on 
$D_{ref}$
is 
susceptible to mode collapse \cite{moFreezeDiscriminatorSimple2020,liFewshotImageGeneration2020}. Hence, as we adapt the model to learn a fairer SA distribution, it is important to preserve the general knowledge efficiently. To this aim, we propose \textbf{ fairTL++} where we include two additional improvements upon fairTL: i) {\em multiple feedback approach}, and ii) {\em Linear-Probing  before Fine-Tuning} (LP-FT).
We find that these two innovations can achieve noticeable gain when applied to fairTL individually. Furthermore, when applied together, we are able to achieve significant gain in sample quality and fairness over previous work \cite{choiFairGenerativeModeling2020a}.
In particular, fairTL and fairTL++ differentiates itself by \textit{removing the need for a density ratio classifier}, which we found to be inaccurate and difficult to train, thereby circumventing the limitations faced in \cite{choiFairGenerativeModeling2020a}.

Next, we take it a step further, and consider an \textbf{\em alternative, challenging problem setup}. In this setup, only pre-trained (potentially biased) models are available, while datasets that were used to pre-train the models are {\em inaccessible}.  
We show that proposed fairTL and fairTL++ methods are also effective under this setup, where they improve both quality and fairness of a pre-trained GAN by adapting it on a small fair dataset. 
We remark that since previous work requires access to the large dataset $D_{bias}$, it is incapable of handling this challenging setup.
The significance of this new setup is that it enables fair and high-quality GANs without imposing access to large datasets and high computational resources. 
\\

Our {\bf main contributions} are:
\begin{itemize}
    \item 
    {In the Choi \etal setup (which assumes availability of both $D_{bias}$ and $D_{ref}$) we show that a simple transfer learning approach --called fairTL-- is very effective for training a fair generative model. We have also proposed fairTL++ by introducing two simple improvements upon fairTL to preserve general knowledge while capturing the fair distribution \wrt SAs during adaptation.}
    \item{ We also introduce a more challenging setup which considers debiasing pre-trained GANs, where only the small fair dataset $D_{ref}$ is available. Both proposed fairTL and fairTL++ approaches remain effective in this setup, paving the way for making better use of pre-trained GANs while addressing fairness.}
    \item {
    We conduct extensive experiments to show that our proposed method can achieve state-of-the-art (SOTA) performance in generated samples quality, diversity and fairness.}
\end{itemize}

\section{Related Work}
{\bf Fairness in Generative Models.}
Fairness in machine learning (ML) is mostly studied for classification problems, where generally the objective is to handle a classification task independent of a SA in the input data, \eg making `hiring' decisions independent of \texttt{Gender}. Different measurement metrics are used for this objective, including well-known Equalised Odds, Equalised Opportunity \cite{hardt2016equality} and Demographic Parity \cite{feldman2015certifying}. 
However, in generative models, fairness is defined as {\em equal representation}, \ie uniform distribution of samples \wrt SAs. 
This results in some misalignment in the objective of fair generative models with earlier classifier works.

Several works have addressed the enforcement of fairness in generative models, often with the use of auxiliary models. Fair-GAN \cite{xu2018fairgan} and FairnessGAN \cite{sattigeri2019fairness} are proposed to generate fair datasets (data-points and labels) as a pre-processing technique. 
In these works, 
a  downstream classifier learns to identify the SA, providing feedback to the generator. 
Nonetheless, all of these works are supervised  
and hence require a large, well-labeled dataset. However in 
the proposed
setup, we do not have access to such a labeled dataset.

Regardless, there exists a few works that adopt a similar unsupervised or semi-supervised approach.
In particular, 
Fair GAN without retraining \cite{tanImprovingFairnessDeep2020}, aims to learn the latent distributions of the input noise \wrt the SA, which then allows us to sample uniformly from it. 
Frankle \etal \cite{frankelFairGenerationPrior2020} introduces the concept of prior modification, where an additional smaller network is added to modify the prior of a GAN to achieve a fairer output. 
Importance weighting algorithm is proposed in \cite{choiFairGenerativeModeling2020a} for the training of a fair generative model. 
In this algorithm, a reference fair dataset \wrt the SA is used during training, while simultaneously exposing the model to the large biased dataset (from which samples are re-weighted). This allows the generator to output high-quality samples, while encouraging fairness \wrt the SA.
SOTA quality and fairness of generated samples has been reported in \cite{choiFairGenerativeModeling2020a}.
Lastly, although not explored deeply, MaGNET \cite{humayunMaGNETUniformSampling2021a} hints at the possibility that enforcing uniformity in the latent feature space of a GAN through a sampling process, may have an impact in enforcing fairness \wrt a SA.

{\bf Transfer Learning.} 
The main idea in transfer learning is to achieve a low generalization risk by adapting a pre-trained model (usually trained on a large-diverse dataset) to a target domain/task 
by 
using usually limited data from 
the same
target domain/task~\cite{pan2009survey, zhao2022fewshot,
congGANMemoryNo2020,zhaoLeveragingPretrainedGANs2020,moFreezeDiscriminatorSimple2020}.
Generally, in {\em discriminative learning}, the pre-trained model 
is adapted in
two simple ways \cite{yosinski2014transferable, jiang2022transferability}: i) {\em Linear-Probing} (LP), which freezes the pre-trained network weights and 
trains the newly added ones 
\cite{wu2020understanding, malekzadeh2017aircraft, du2020few}, and ii) {\em fine-tuning} (FT) which continues to train using the entire pre-trained network weights \cite{cai2019device, guo2019spottune, abdollahzadeh2021revisit}.
Recently, \cite{kumarFineTuningCanDistort2022} suggests that utilizing  Linear-Probing prior to Fine-Tuning (LP-FT) can help preserve important features needed for adaptation.
In {\em generative learning}, TGAN \cite{wangTransferringGANsGenerating2018} demonstrates the effectiveness of transferring pre-trained GANs into new domains, thereby improving performance with limited data.
CDC \cite{ojhaFewshotImageGeneration2021} uses a similar approach in Few-shot Cross-domain Adaptation,
but with the addition of a cross-domain consistency loss.
EWC \cite{liFewshotImageGeneration2020} 
 discusses the preservation of certain weights during adaptation to maintain the diversity of the source domain.
In contrast to the previous works that aims to address 
the improvement of sample quality
on the target domain, we address a  different concept --{\em improving the fairness} using transfer learning.

\textbf{Multiple Feedback Approach}. Learning through a multiple feedback approach has been a popular approach in improving quality of generated samples, particularly when faced with limited samples
\cite{kumariEnsemblingOfftheShelfModels2022,tranDataAugmentationGAN2021}. 
Instead of the standard one-generator-and-discriminator approach, the multiple feedback approach takes advantage of multiple discriminators \cite{NIPS2017_6860,durugkarGENERATIVEMULTIADVERSARIALNETWORKS2017,albuquerqueMultiobjectiveTrainingGenerative2022,umFairGenerativeModel2021} or multiple generators, thereby improving stability during optimization \cite{hoangMGANTrainingGenerative2022,ghoshMultiAgentDiverseGenerative2018}.

\section{Proposed Method}
In this section, we first consider the problem setup in \cite{choiFairGenerativeModeling2020a} -- which assumes the availability of $D_{bias}$ and $D_{ref}$ -- and outline the details of the proposed fairTL and its improved variant fairTL++.
Next, we describe a new challenging problem setup that removes the need for a large biased dataset $D_{bias}$, and only considers the availability of a pre-trained (possibly biased) GAN and a small fair dataset $D_{ref}$. Existing methods can not handle this setup because of their reliance on $D_{bias}$ for training a fair GAN.

\subsection{fairTL}
Here, we present a simple transfer learning-based method in training a GAN for fair, diverse and high-quality sample generation, based on  
$D_{bias}$ and $D_{ref}$. 
This process includes a pre-training step, which is followed by adaptation.
In the pre-training stage, we train the generative model to learn the required general knowledge for sample generation, 
using all available training data.
In particular, we train the model with GAN loss  \cite{goodfellow2014GAN}. 
We remark that our approach is not restricted to a particular loss function.
Here, we define $\G_s$ and $\D_s$ as the biased generator and discriminator in the 
pre-training stage, trained on samples in $D_{bias} \cup D_{ref}$. 
Next, in the adaptation stage, using the same loss function, we adapt the generative model
to learn fair SA distribution by using $D_{ref}$ only:
\begin{equation}
    \begin{split}
    \label{pretrain}
    \min_{\G_t}\max_{\D_t}\mathcal{L} = \mathbb{E}_{x \in  D_{ref} }[\log \D_t(x)] \\
    + \mathbb{E}_{z\sim p_{z}(z)}[\log{(1-\D_t(\G_t(z)))}].
    \end{split}
\end{equation}
Here,  $\G_t$, $\D_t$ are  generators and discriminators in the 
adaptation stage, trained on samples {\em only} in  $D_{ref}$, and $z$ is random noise sampled from a Gaussian noise distribution $p_z(z)$. 
Furthermore, $\G_t$, $\D_t$ are initialized from $\G_s$, $\D_s$ respectively.
Our experimental results show that this simple approach can be considered as a {\em strong baseline} for fair GAN training which achieves competitive performance with the SOTA method proposed in \cite{choiFairGenerativeModeling2020a}.

\subsection{fairTL++}

One technical challenge of using fairTL is that due to the small size of $D_{ref}$,
fine-tuning on 
$D_{ref}$ is susceptible to mode collapse \cite{moFreezeDiscriminatorSimple2020,liFewshotImageGeneration2020}.
To prevent the model from forgetting the general knowledge learned during pre-training, we propose \textit{fairTL++}, which includes two additions when adapting to $D_{ref}$: Linear-Probing before Fine-Tuning (LP-FT), and a multiple feedback approach during adaptation (Figure. \ref{fig:SampleIllustration}). In what follows, we discuss the details of each method.

\textbf{a) LP-FT.} 
\cite{kumarFineTuningCanDistort2022} demonstrates that when adapting a pre-trained classifier to a new task, it is advantageous to first use Linear-Probing (updating the classifier head but freezing lower layers) for some limited epochs $T$, and then use Fine-Tuning (updating all model parameters). This method is termed LP-FT.
Experimental results in \cite{kumarFineTuningCanDistort2022} suggests that Linear-Probing allows for more task-specific parameters to adapt before Fine-Tuning, and generally works better for transfer learning.
We found that a similar approach can be adopted for our generative learning setup. 
In our context, the discriminator can be considered as the feature extractor, and the downstream task is to learn the fairer SA distribution of $D_{ref}$.

To implement this, we first conduct an empirical study to identify the SA-specific layers needed for adaptation.
In this study, we similarly implement fairness adaptation with fairTL, but with a large $D_{ref}$, thereby alleviating the instability during training. 
Next, we evaluated the mean layer weight change.
In our results, we observed that amongst 
the layers in $\D_t$ and $\G_t$
only the first two layers of $\D_t$ (closest to the model's input) expressed low changes in their weight, 
thereby indicating that they 
are the least associated with the SA. Hence, these are general layers that should be preserved.
To validate this, we implemented LP 
while freezing different layer permutations
and similarly found that freezing any additional layers, other than the first two layers of $\D_t$, resulted in 
poorer sample quality. We found that these results were consistent across several different SA.
This finding aligns with works from domain adaptation \cite{moFreezeDiscriminatorSimple2020}, which similarly found it advantageous to retain (freeze) the lower-level layers of the discriminator \textit{throughout} 
fine-tuning. 
However, we noted that retaining the low-level layers throughout the adaptation stage creates instability \ie the generator start to output noise. Conversely, retaining those same layers for only $T$ epoch improves quality and fairness. 
Therefore, when adapting $\G_s$ and $\D_s$ into fair dataset $D_{ref}$, we first freeze the lower layers of the discriminator for some limited epochs, and then fine-tune all parameters.



\begin{table*}[h!]
    \centering
    \resizebox{\textwidth}{!}{
    \begin{tabular}{
    c @{\hspace{1em}} 
    c @{\hspace{1em}} c @{\hspace{1em}} c @{\hspace{1em}} c | 
    c @{\hspace{1em}} c @{\hspace{1em}} c @{\hspace{1em}} c @{\hspace{1em}} 
    }
    \toprule
    & \multicolumn{4}{c}{90\_10} & \multicolumn{4}{c}{Multi}  \\
    \cmidrule(lr){2-5} \cmidrule(lr){6-9}
    Perc    &  0.25 &  0.1 &  0.05 &  0.025
            &  0.25 &  0.1 &  0.05 &  0.025\\ 
    \midrule
    \multicolumn{9}{c}{a) Imp-weighting \cite{choiFairGenerativeModeling2020a}}\\
    \midrule
    FID($\downarrow$) & $19.20 \pm 0.10$  & $20.42 \pm 0.20$ & $23.01 \pm 0.15$  & $25.82 \pm 0.13$ 
                      & $14.61  \pm 0.21$ & $16.92 \pm 0.31$ & $19.43 \pm 0.23$  & $22.80  \pm 0.13$\\
    FD ($\downarrow$) & $0.090\pm 0.011$  & $0.107 \pm 0.022$ & $0.167 \pm 0.016$ & $0.246\pm 0.032$
                     & $0.142 \pm 0.032$  & $0.116 \pm 0.020$ & $0.135 \pm 0.014$ & $0.144 \pm 0.016$\\
    \midrule
    \multicolumn{9}{c}{b) fairTL}\\
    \midrule
    FID($\downarrow$) & $\underline{14.21} \pm 0.02$ & $ \underline{20.00} \pm 0.10$  & $ \underline{22.99} \pm 0.09$ & $ \underline{23.60} \pm 0.11$
                      & $\underline{11.98} \pm 0.12$ & $\underline{13.10}  \pm 0.14$  & $\underline{13.29} \pm 0.16$  & $\underline{13.99} \pm 0.10$\\
    FD ($\downarrow$) & $\underline{0.087} \pm 0.007$ & $\underline{0.105} \pm 0.020$  & $ \underline{0.107} \pm 0.012$ & $ \underline{0.130} \pm 0.029$
                      & $\underline{0.113} \pm 0.021$ & $\underline{0.115} \pm 0.017$  & $\underline{0.118} \pm 0.013$ & $\underline{0.138} \pm 0.011$\\
    \midrule
    \multicolumn{9}{c}{c) fairTL++}\\
    \midrule
    FID($\downarrow$) & ${\bf 9.02  \pm 0.03}$  & {$\bf 10.69 \pm 0.11$} & $\bf{20.12 \pm 0.04}$ & $\bf{20.70 \pm 0.08}$
                      & $\bf{10.50 \pm 0.10}$   & $\bf{11.38 \pm 0.11}$  & $\bf{12.00 \pm 0.10}$  & $\bf{13.18 \pm 0.06}$ \\
    FD ($\downarrow$) & ${\bf 0.010 \pm 0.007}$  & {$\bf 0.062 \pm 0.022$} & $\bf{0.035 \pm 0.034}$ & $\bf{0.092\pm 0.025}$
                      & $\bf{0.016 \pm 0.010 }$ & $\bf{0.090\pm 0.020}$   & $\bf{0.086 \pm 0.020}$  & $\bf{0.101 \pm 0.016}$ \\
    \bottomrule
    \end{tabular}
    }
    \caption{\textbf{Comparing our proposed Fair Transfer Learning against Imp-weighting \cite{choiFairGenerativeModeling2020a} on CelebA \cite{liu2015faceattributes}, for single SA (\texttt{Gender}) and multi-SA (\texttt{\{Gender,Blackhair\}}).
    } 
    For \textit{single SA} (\texttt{Gender}), we utilize a $D_{bias}$ with $bias=90\_10$ \ie 90\% sample are Female and 10\% Male, 
    and
    for \textit{multi-SA} a $D_{bias}$ with bias F-NBH, F-BH, M-NBH, M-BH=$[0.437,0.063,0.415,0.085]$ (\texttt{Male}(M), \texttt{Female}(F), \texttt{BlackHair}(BH) and \texttt{No-BlackHair}(NBH)).
    Then, we varied the sample size of 
    $D_{ref}$, while $|D_{bias}|$ is kept constant. This is denoted by the ratio $perc=|D_{ref}|$/$|D_{bias}|$ for $\{0.25, 0.1, 0.05, 0.025\}$. 
    With this setup, we utilize BIGGAN \cite{brockLargeScaleGAN2019} to reproduce (a) \cite{choiFairGenerativeModeling2020a} the current SOTA results,
    and implement our proposed (b) fairTL and (C) fairTL++. We show that our proposed method fairTL is effective in achieving new SOTA FID and FD results for all $perc$, while fairTL++ demonstrates even greater improvements.
    For FID and FD, a low score indicates higher quality samples and fairer SA distribution, respectively.}
    \label{tab:BIGGANCelebABigTable}
\end{table*}

\begin{table}[h!]
    \centering
    \resizebox{0.9\columnwidth}{!}{
    \begin{tabular}{
    c @{\hspace{3em}} 
    c @{\hspace{3em}} 
    c @{\hspace{3em}} c @{\hspace{2em}}
    }
    \toprule
    Perc    &  0.25 &  0.1 \\ 
    \midrule
        \multicolumn{3}{c}{a) Imp-weighting \cite{choiFairGenerativeModeling2020a}}\\
    \midrule
    FID($\downarrow$) & $27.57 \pm 0.45$ & $39.03 \pm 0.72$\\
    FD ($\downarrow$) & $0.154 \pm 0.031$ & $0.205 \pm 0.044$\\
    \midrule
            \multicolumn{3}{c}{b) fairTL}\\
    \midrule
    FID($\downarrow$) & $\underline{20.70} \pm 0.32$ & $\underline{22.92} \pm 0.22$ \\
    FD ($\downarrow$) & $\underline{0.044} \pm 0.017$ & $\underline{0.039} \pm 0.015$\\
    \midrule
            \multicolumn{3}{c}{c) fairTL++}\\
    \midrule
    FID($\downarrow$) & {$\bf 19.21 \pm 0.32$} & {$\bf 21.22 \pm 0.19$}\\
     FD ($\downarrow$) & {$\bf 0.018 \pm 0.020$}& {$\bf 0.003 \pm 0.002$}\\
    \bottomrule
    \end{tabular}
    }
    \caption{ \textbf{Comparing our proposed Fair Transfer Learning against Imp-weighting \cite{choiFairGenerativeModeling2020a} on UTKFace \cite{zhifei2017cvpr}, for \textit{single SA} (\texttt{Race-Caucasian})
    .} 
    We utilize the same \textit{single SA} setup, as per Tab. \ref{tab:BIGGANCelebABigTable}.
    Given that UTKFace is a small dataset, we are limited to $perc=\{0.25,0.1\}$.
    We then similarly utilize BIGGAN and compare a) \cite{choiFairGenerativeModeling2020a} as the current SOTA against our proposed b) fairTL and c) fairTL++. Similarly, we find that our proposed solutions outperform Choi \etal in both FID and FD.}
    \label{tab:BIGGANUTKFace}

\end{table}

\textbf{ b) Multiple Feedback.} 
\cite{kumariEnsemblingOfftheShelfModels2022,umFairGenerativeModel2021} proposes that the utilization of collective knowledge from multiple pre-trained discriminators improves GAN performance under limited data settings.
Inspired by this, we consider that our pre-trained discriminator $\D_s$ is proficient at evaluating the quality of our generated samples despite being trained on a biased dataset.    
With this,
we adopted a multiple feedback approach during our adaptation
stage. 
In particular, we retain a \textit{frozen} copy of our discriminator $\D_{s}$ after pre-training and append it to our model. Similarly, we carry out adaptation on $D_{ref}$ with $\D_s$, $\D_t$ and $\G_t$. During this process, only $\D_t$ and $\G_t$ weights are updated. Intuitively, $\D_s$ can be seen to discriminate upon the generated sample quality, while the $\D_t$ adapts to the $D_{ref}$ and enforces the new fair SA distribution. 
Eqn. \ref{eqn:fairTL++} presents the loss function, where we utilize $\lambda \in [0,1]$ as a hyper-parameter to control the balance between enforcing fairness and quality. 
In our experiments, we found that although
both discriminators play an essential part in improving the performance of the GAN, more emphasis should be placed on $\D_t$.
In particular, since $\D_s$ is frozen, making $\lambda$ too small results in instability during training. Conversely, making $\lambda$ too big limits the feedback we get on the sample's quality. 
Empirically, we found $\lambda=0.6$ to be ideal.

\begin{align}
    \begin{split}
    \label{eqn:fairTL++}
    \min_{\G_t}\max_{\D_t}\mathcal{L} = \mathbb{E}_{x \in  D_{ref} }[\log \D_t(x)] \\
    + \lambda\mathbb{E}_{z\sim p_{z}(z)}[\log{(1-\D_t(\G_t(z)))}]\\
    + (1-\lambda)\mathbb{E}_{z\sim p_{z}(z)}[\log{(1-\D_s(\G_t(z)))}]
    \end{split}
\end{align}

As we will later discuss in our experiments, having the addition of LP-FT and multiple feedback approach improves the stability of our training process, and allows our proposed method to achieve SOTA quality and fairness. 

\subsection{Improving the Fairness of Pre-trained GANs}
As mentioned before, when discussing our proposed fairTL and fairTL++, we assume that similar to \cite{choiFairGenerativeModeling2020a}, we have access to a large biased dataset $D_{bias}$, and a small fair dataset $D_{fair}$.
Under this setup, the generative model requires to be trained from scratch, which entails significant computational resources.
Also, a large dataset is necessary to have expressive representation.
Another solution to learn a fair generative model, which can output diverse and high-quality samples, is to take advantage of (potentially biased) pre-trained generative models, and improve their fairness \wrt the desired SAs.
Under this new challenging setup, we assume that there is a pre-trained GAN, but the dataset used for pre-training is {\em inaccessible}. However, we have access to a small, fair dataset from the related distribution.
Since our proposed fairTL and fairTL++ methods are based on the general idea of transfer learning, they can be easily adapted to this challenging setup by discarding the first pre-training step.
Our experimental results show that fairTL and fairTL++ remain effective under this setup, and can improve the fairness and quality of SOTA pre-trained GANs.


\begin{table*}[h!]
    \centering
    \resizebox{0.8\textwidth}{!}{
    \begin{tabular}{
    c @{\hspace{1em}} 
    c @{\hspace{1em}} c @{\hspace{1em}} c @{\hspace{1em}} c @{\hspace{1em}} c @{\hspace{1em}} 
    }
    \toprule
    Perc                    & \multicolumn{5}{c}{0.025} \\
    \cmidrule(lr){2-6}
    Sensitive Attributes    & \texttt{Gender} &  \texttt{BlackHair} &  \texttt{Young} &  \texttt{Smiling} & \texttt{Moustache}  \\ 
    \midrule
    \multicolumn{6}{c}{a) Pre-trained}\\
    \midrule
    FID($\downarrow$) & $9.20 \pm 0.02$  & $ 14.58 \pm 0.11$  & $ 24.60 \pm 0.21$  & $9.30 \pm 0.03$  & $19.84\pm 0.21$ \\
    FD ($\downarrow$) & $0.102\pm 0.019$  & $ 0.075 \pm 0.002$ & $ 0.277 \pm 0.012$   & $0.168 \pm 0.007$ & $0.376\pm 0.041$ \\
    \midrule
        \multicolumn{6}{c}{b) fairTL}\\
    \midrule
    FID($\downarrow$) & $\underline{9.01 \pm 0.01}$  & $\underline{13.39 \pm 0.09}$  & $\underline{12.94 \pm 0.10}$ & $\underline{9.15 \pm 0.01}$  & $\underline{13.03 \pm 0.14}$\\
    FD ($\downarrow$) & $\underline{0.088 \pm 0.010}$ & $\underline{0.058 \pm 0.012}$ & $\underline{0.093 \pm 0.023}$ & $\underline{0.098 \pm 0.020}$ & $\underline{0.096 \pm 0.042}$\\
    \midrule
            \multicolumn{6}{c}{c) fairTL++}\\
    \midrule
    FID ($\downarrow$) &  ${\bf 8.81 \pm 0.01}$  & ${\bf 12.32 \pm 0.10}$ & ${\bf 11.79 \pm 0.12}$ & ${\bf 8.90\pm 0.02}$  & ${\bf 11.66 \pm 0.14}$ \\
    FD  ($\downarrow$) &  ${\bf 0.067 \pm 0.014}$ & ${\bf 0.057 \pm 0.008}$ & ${\bf 0.056 \pm 0.011}$ & ${\bf 0.061 \pm 0.023}$ & ${\bf 0.025 \pm 0.028}$\\
    \bottomrule
    \end{tabular}
    }
    \caption{\textbf{Evaluating our proposed Fair Transfer Learning on a pre-trained
    StyleGAN2 \cite{karrasAnalyzingImprovingImage2020} and FFHQ \cite{karrasStyleBasedGeneratorArchitecture2019} dataset.}
    We evaluate our proposed method on the SA \texttt{\{Gender, Blackhair, Young, Smiling, Moustache\}}. 
    As our baseline, we first evaluate the a) Pre-trained model's FID and Fairness (FD) \wrt the different SA.
    Then, utilising a $perc=|D_{ref}|/|D_{bias}|$ of 
    0.025
    , we implement b) fairTL and c) fairTL++ and similarly measure the debiased StyleGAN2 FID and FD.
    Based on our results, we demonstrate that our proposed method is advantageous across SA in improving diversity, quality and fairness of the generated samples \wrt the SA.} 
    \label{tab:StyleGAN20.02}
\end{table*}
\section{Experiments}
\label{sec:experiments}

In this section, we evaluate the performance of the proposed fairTL and fairTL++ in two different problem setups: 1) problem setup of \cite{choiFairGenerativeModeling2020a} where both $D_{bias}$ and $D_{ref}$ are available for a given SA, 2) the proposed problem setup in this work where we have access to only the small $D_{ref}$ and a pre-trained GAN in-place of $D_{bias}$.

For the first setup, we compare our proposed method against importance weighting \cite{choiFairGenerativeModeling2020a} which produces SOTA in quality and fairness.
As importance weighting \cite{choiFairGenerativeModeling2020a} cannot be applied to the second setup due to the unavailability of the large dataset $D_{bias}$, 
we evaluate the performance of the proposed method on mitigating the (potential) existing bias in SOTA pre-trained GANs.
We remark that in both setups, {\em none of the fairness enforcement methods have access to the labels of the datasets} and that these labels are only used
as a controlled means to re-sample the respective datasets and simulate the bias.

{\bf Evaluation Metric.}
Following \cite{choiFairGenerativeModeling2020a}, we utilize FID \cite{heuselGANsTrainedTwo2018a} to evaluate the quality and diversity of our generated samples, and the fairness discrepancy metric (FD) \cite{choiFairGenerativeModeling2020a} to measure the fairness of our models \wrt a SA. 
Similar to \cite{choiFairGenerativeModeling2020a}, when evaluating FID, we re-sample the original large dataset (\eg CelebA) to obtain equal SA representation, which we use to calculate the reference statistics. 
This is necessary as it
allows us to obtain an estimate of the quality and diversity of the generator, 
while referencing 
our target ideal generator
with
fair SA distribution.
Then, to evaluate fairness, we train a ResNet-18 \cite{heDeepResidualLearning2016} 
to classify the generated samples SA, which we use to calculate FD as follows:

\begin{equation}
f  
= |\bar{p} - \mathbb{E}_{z\sim p_{z}(z)} [C(\mathbf{\G(z)})] |_2
    \label{eqn:FD}
\end{equation}
Here, $C(\G(z))$ is the one-hot vector for the classified label of the generated sample $\G(z)$, whose generator can either be $\G_t$ or $\G_s$ depending on the method used. $z$ is sampled from a Gaussian noise distribution $p_z(z)$ and $\bar{p}$ is a uniformly distributed vector with the same cardinality as $C(\mathbf{\G(z)})$. 

\subsection{Setup 1: Training a Fair Generator}
Utilising the setup from \cite{choiFairGenerativeModeling2020a}, we implement our proposed method, by first training a BIGGAN \cite{brockLargeScaleGAN2019} model with all the available data ($D_{bias} \cup D_{ref}$)
to achieve the highest quality generator.
This is then followed by the adaptation stage, with $D_{ref}$ only. 
For a fair comparison, we utilize the source code from \cite{choiFairGenerativeModeling2020a} to reproduce their proposed importance-weighting (imp-weighting) on BIGGAN.

{\bf Dataset.} We consider the datasets CelebA \cite{liu2015faceattributes} and UTKFace \cite{zhifei2017cvpr} for this experiment. For CelebA, following Choi \textit{et al.}, we utilize the SA \texttt{Gender} and \{\texttt{Gender},\texttt{Blackhair}\} for both single and multi-attribute settings respectively. 
Then, for UTKFace, we utilize the SA \texttt{Race(Caucasian)}.
In both single attribute settings, we synthetically introduce a $bias=0.9$ to $D_{bias}$ \ie $D_{bias}$ contains 90\% Female/Caucasian samples and 10\% Male/Non-Caucasian samples, by re-sampling the dataset. For multi-attribute settings, given the data limitations, we similarly introduce a  $D_{bias }$ through re-sampling with the following sample ratios F-NBH, F-BH, M-NBH, M-BH= $[0.437,0.063,0.415,0.085]$ for \texttt{Male}(M), \texttt{Female}(F), \texttt{Blackhair}(BH) and \texttt{no-Blackhair}(NBH). 
Next,
we considered different 
$|D_{ref}|$ while keeping $|D_{bias}|$ constant,
 notated by $perc=|D_{ref}|$/$|D_{bias}|$.
 This allows us to evaluate the robustness of our proposed method with decreasing reference samples during adaptation.
 For the CelebA dataset, we explore $perc=\{0.25,0.1,0.05,0.025\}$ and for UTKFace, due to its smaller dataset, we explore $perc=\{0.25,0.1\}$.

{\bf Single Attribute Results.}
Table \ref{tab:BIGGANCelebABigTable} presents the results of imp-weighting \cite{choiFairGenerativeModeling2020a} against our proposed methods on the CelebA dataset. 
Comparing across 
different $perc$,
we observe that fairTL is generally able to outperform imp-weighting, achieving better fairness and quality. 
Then, with the addition of LP-FT and the multi-feedback approach, we observed greater improvements in fairTL++,
highlighting the effectiveness of these two additions during adaptation.
In particular, we notice that even with the smallest reference dataset, $perc=0.025$, fairTL++ is able to achieve a relatively fair generator, \ie low FD measurement, while imp-weighting worsens under these conditions.
Table \ref{tab:BIGGANUTKFace} then compares the same methods but on the UTKFace dataset with SA \texttt{Race-Caucasian}. 
With this dataset, we similarly observe that fairTL++ outperforms both imp-weighting and fairTL in quality and fairness. 
In fact, on the smaller UTKFace dataset, the benefits of our proposed method becomes more prominent with imp-weighting's sample quality (FID) significantly degrading as $perc$ becomes smaller. In contrast, our proposed method only experiences minor degradation, while still enforcing SOTA fairness (FD).

{\bf Multi-attribute results.} Table \ref{tab:BIGGANCelebABigTable} presents a similar experiment but with multiple SA \texttt{Gender} and \texttt{Blackhair}. Our results show that even under this more challenging setup, involving two SA simultaneously, fairTL++ still outperforms imp-weighting, thereby achieving SOTA performance in mitigating bias while maintaining high-quality samples.

\begin{figure}[ht]
    \centering
    \includegraphics[width=\columnwidth]{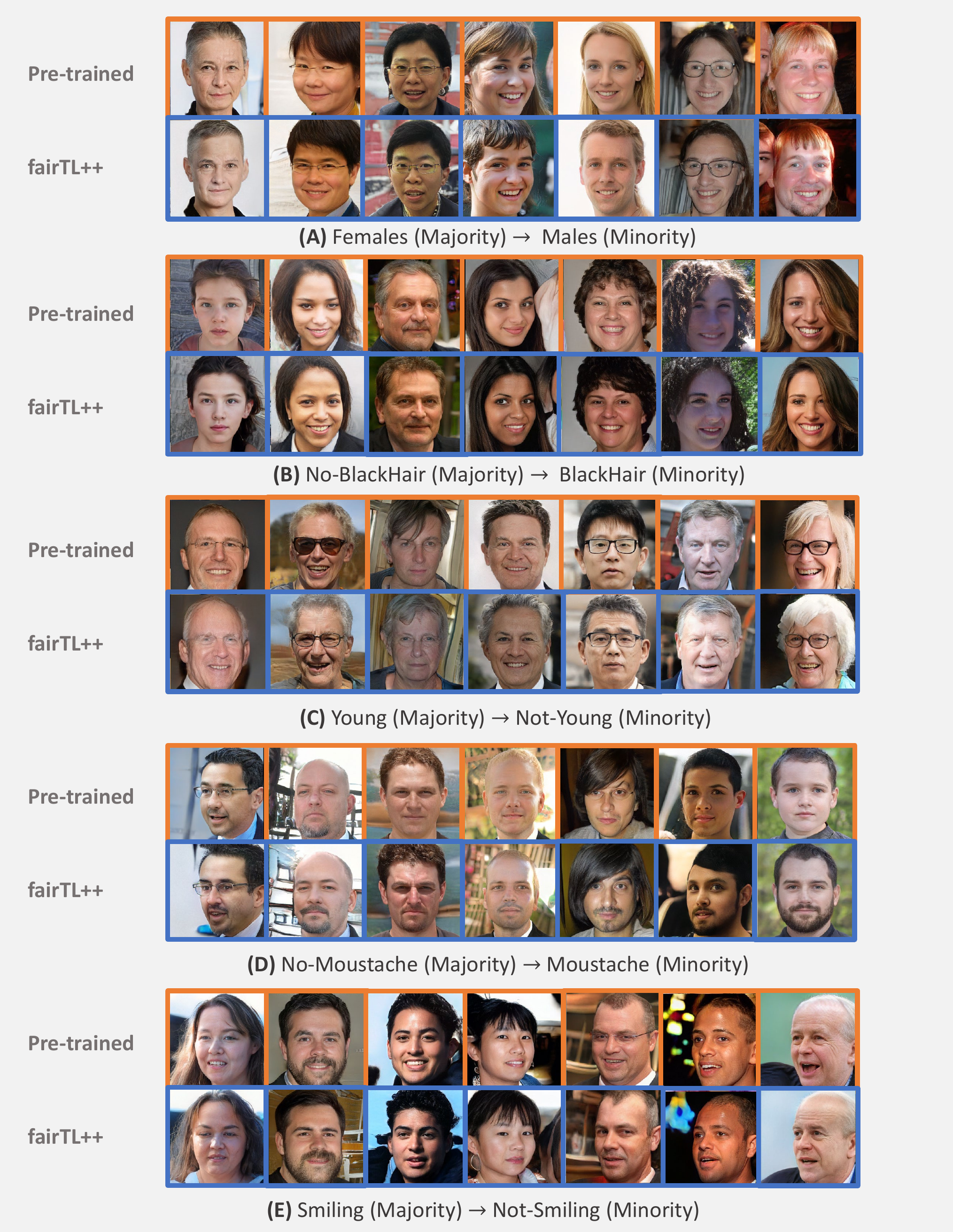}
    \caption{{\bf Illustration of samples before and after fairness adaptation by our fairTL++ on a pre-trained StyleGAN2 \cite{karrasAnalyzingImprovingImage2020}.} For each sample, we utilize the same noise vector to sample from the pre-trained model and fairTL++ after SA adaptation. Notice how the samples are adapted from the majority to minority represented SA.} 
    \label{fig:StyleGAN2}
\end{figure}
\subsection{Setup 2: Debiasing a Pre-trained Generator}
In this new setup, we demonstrate that unlike previous works, our proposed method does not strictly require access to the large dataset ($D_{bias}$).
Instead, we are able to improve on the fairness of existing biased pre-trained models. 
For this experiment, we utilize the original code as in StyleGAN2 \cite{karrasAnalyzingImprovingImage2020} as the baseline, along with the pre-trained weights on the FFHQ dataset \cite{karrasStyleBasedGeneratorArchitecture2019}. 
With this baseline, 
we followed the same setup as the previous experiments for fair comparison, and 
measured the FID and FD of the pre-trained model across different SA. Then utilizing $D_{ref}$ we implement the adaptation stage for fairTL and fairTL++ and re-evaluated the model.

{\bf Dataset.} We utilize the FFHQ dataset and consider the SA \{\texttt{Gender}, \texttt{Blackhair}, \texttt{Young}, \texttt{Smiling},  \texttt{Moustache}\} to demonstrate the effectiveness of our proposed method across different SA. For each SA, we attained a $D_{ref}$
with $perc$=0.025.


{\bf From our results} in Table \ref{tab:StyleGAN20.02}, we observe that the pre-trained StyleGAN2 model contains a considerable amount of bias in the selected SA. In particular, larger biases exist for SA \{\texttt{Young,Smiling,Moustache}\} 
, where high FD measurements were reported.
Furthermore, the high FID measurements indicates a mismatch between the the diversity of the generated samples and the ideal reference samples.
Our proposed solutions however proves to be effective in improving both fairness and diversity of the StyleGAN2, while achieving high-quality samples, as seen from the relatively low FD and FID score.
Similar to the previous experiments, fairTL++ proves to be the more effective method.
Fig. \ref{fig:StyleGAN2} illustrates a few samples that have been adapted from the majority-represented SA to the minority-represented SA, thereby achieving a fairer SA distribution.
We remark that though the SA of the samples have been adapted \eg \texttt{Female} to \texttt{Male}, the underlying general attribute \eg pose and race remain unchanged.
\section{Conclusion}
In our work, we focus on the challenging task of training a diverse, high-quality GAN while achieving fairness \wrt some sensitive attributes. In this task, we are given the real world constraints of having only access to a small but fair dataset and a large but biased dataset. To overcome these limitations, we propose a simple and effective method of
training a fair generative model via transfer learning. To do this, we first pre-train the model with the large biased dataset, followed by  fairness adaptation with the small unbiased dataset. 
We then further demonstrate that the introduction of a multiple feedback approach and Linear-Probing to the sensitive attribute specific layers during adaptation, can help further improve both sample quality and fairness, thereby achieving state-of-the-art performance. 
Additionally, we demonstrate that our proposed methods can similarly improve the quality and fairness of SOTA pre-trained GANs.

\section{Acknowledgement}
This research is supported by the National Research Foundation, Singapore under its AI Singapore
Programmes (AISG Award No.: AISG2-RP-2021-021; AISG Award No.: AISG2-TC-2022-007).
This project is also supported by SUTD project PIE-SGP-AI-2018-01. We thank anonymous reviewers
for their insightful comments.

\bibliography{main.bib,TIPGAN.bib} 
\end{document}